\documentclass[letterpaper, 10 pt, conference]{ieeeconf}  

\IEEEoverridecommandlockouts                              

\usepackage{amsmath}
\usepackage{url}
\usepackage{booktabs} 
\usepackage{array} 
\usepackage{graphicx}
\usepackage{verbatim} 
\usepackage{forest}

\title{\LARGE \bf
	Low-viewpoint forest depth dataset for sparse rover swarms}

\author{Chaoyue Niu, Danesh Tarapore and Klaus-Peter Zauner
  \thanks{School of Electronics and Computer Science, University of Southampton, Southampton, U.K.}
  \thanks{Corresponding author: Chaoyue Niu {\tt\small cn1n18@soton.ac.uk}}
}

\begin{document}

\maketitle
\thispagestyle{empty}
\pagestyle{empty}

\begin{abstract}
  Rapid progress in embedded computing hardware increasingly enables
  on-board image processing on small robots. This development opens
  the path to replacing costly sensors with sophisticated computer
  vision techniques. A case in point is the prediction of scene depth
  information from a monocular camera for autonomous navigation.
  Motivated by the aim to develop a robot swarm suitable for sensing,
  monitoring, and search applications in forests, we have collected a
  set of RGB images and corresponding depth maps. Over 100000 RGB/depth image pairs
  were recorded with a custom rig from the perspective of a small
  ground rover moving through a forest. Taken under different
  weather and lighting conditions, the images include scenes with grass, bushes,
  standing and fallen trees, tree branches, leaves, and dirt. In
  addition GPS, IMU, and wheel encoder data were recorded. From the
  calibrated, synchronized, aligned and timestamped frames about
  9700 image-depth map pairs were selected for sharpness and variety.
  We provide this dataset to the community to fill a need identified
  in our own research and hope it will accelerate progress in robots
  navigating the challenging forest environment. This paper
  describes our custom hardware and methodology to collect the data,
  subsequent processing and quality of the data, and how to
  access it.
\end{abstract}


\section{Introduction}

Forests are ecologically and economically important, affecting the
local as well as wider climate and are under pressure from changing
weather patterns and deceases. They are also a formidable challenge
for small all-terrain ground robots. In ongoing research we are aiming
at developing a rover platform for this environment. We
envisage robot swarms as a useful tool in the efforts to protect,
reform, and extend forests.

Robots swarms are teams of robots that coordinate their actions in a distributed fashion to perform an assigned task. A common feature of existing swarms is the underlying assumption that the robots act in close proximity to each other
\cite{brambilla2013swarm}. For real-world, outdoor applications over
extended areas, such a density is neither desirable nor feasible. A
dense swarm would not only be very costly, but also highly intrusive to the environment.
Recently available technologies in long range radio communication and
efficient battery technologies, however, allow for the
reconceptualisation  of swarms as scalable groups of robots acting
jointly over distances up to 1~km. Such robots need to be low cost
and high in autonomy.

Safely navigating mobile robots in off-road environments such as in a forest, requires real-time and accurate terrain traversability analysis.
To enable safe autonomous operation of a swarm of robots
during exploration, the ability to accurately estimate terrain
traversability is critical.
By analyzing geometric features such as the depth map or point cloud, and appearance characteristics such as colour or texture, a terrain can be analysed with respect to the mechanical and locomotion constraints of a robot \cite{balta2013terrain}.
To support this analysis for off-road path planning we are developing
a vision system required to run on small, on-board computers. To also
keep the cost of sensors low, we are interested in monocular depth
estimation \cite{bhoi2019monocular} to predict local depth from single
images or sequences of images from a single moving camera. Aside from
optical flow \cite{ho2017distance} and geometric techniques \cite{hartley2003multiple, oram2001rectification}, machine learning has been
applied to achieve this. A number of authors have trained depth
estimation models by using deep neural network architectures
(\cite{godard2017unsupervised,xu2018structured,eigen2014depth,laina2016deeper,alhashim2018high}).

Most existing outdoor depth map datasets focus on unmanned driving
applications. The KITTI dataset \cite{geiger2013vision} records street
scenes in cities. The Freiburg Forest dataset \cite{valada16iser}
records the forest view from a cart track and lacks a close-range
perspective. Because this dataset was manually labeled for image
segmentation it is comprised of only 366 images and therefore too
small to train deep neural networks. The Make-3D dataset
(\cite{saxena2008make3d,saxena2007learning}) records outdoor
scenes including some with close-up depth data, but it mainly concentrates on
buildings in a city. We have found that most of the publicly
available RGB-D datasets are recorded indoors \cite{firman2016rgbd}.
While close-range depth data is available for these indoor conditions
\cite{Silberman:ECCV12}, this was so far not the case for
natural outdoor environments. Accordingly, the available depth
datasets were not suitable for our purpose.
Moreover, a common feature of the above datasets is that the images are taken
from a high point of view. Our interest is in small portable robots
that can be transported with a backpack. The camera perspective of these
robots will be from a low viewpoint and we therefore prefer a depth
dataset with such a low perspective.


\section{Mobile sensor platform setup}
\begin{figure}
  \centering
  \includegraphics[width=3in]{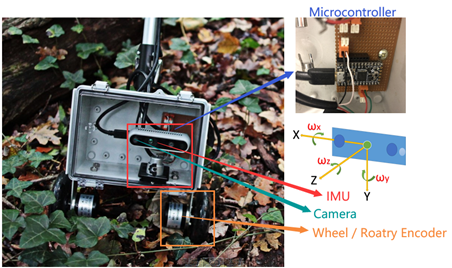}
  \caption{\textbf{Depth data collection rig.} The recording system is
    equipped with an Intel D435i global shutter depth camera, two
    rotary encoders, and a GPS. A microcontroller monitors the
    incremental rotary encoders and interfaces them to a USB
    connection.}
  \label{fig:platform}
\end{figure}

\begin{figure}
	\centering
	\includegraphics[width=3in]{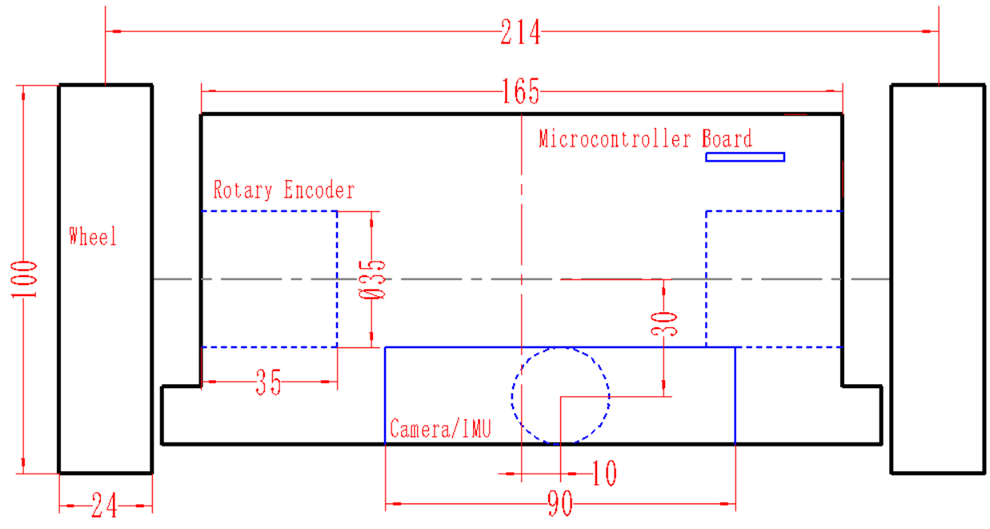}
	\caption{\textbf{Sensor configuration.} Top view of the
          mounting positions and dimensions of the sensors on
          the depth data collection rig. Solid black lines represent
          the wheels and the box; blue lines represent the sensors.
          Dimensions in millimeter; the camera lens is located 150~mm above ground.}
	\label{fig:setup}
\end{figure}

To facilitate efficient data collection we decided to manually move
the camera along the path to be recorded, rather than to
record with a robot-mounted camera. The recording rig shown in
Fig.~\ref{fig:platform} was constructed by attaching two
incremental photoelectric rotary encoders to an electrical enclosure
box and mounting a 100~mm diameter wheel to each encoder. The encoders
were connected to a Micropython enabled ARM board (ItsyBitsy M4
Express, Adafruit, NY, USA.) which made the time stamped rotary encoder
readings available over a USB connection.
The enclosure was mounted at the end of a telescopic rod of the type used for
paint rollers. This allows the user to roll the enclosure on its
wheels along the ground by pushing it forward while walking.
Inside the enclosure a RealSense D435i depth camera (Intel, CA, USA) was
mounted 150~mm above ground with a free field of view in the direction of motion as
illustrated in Fig.~\ref{fig:setup}.
The D435i camera combines a depth camera and an RGB colour camera with
a diagonal field of view of 94$^{\circ}$ and 77$^{\circ}$, respectively. With its global shutter, this camera is well
suited to a moving platform, and it also contains an
inertial measurement unit (IMU). A laptop computer is connected to the
camera, to the USB connection from the rotary encoders and
to a GPS receiver (BU-353-S4 SiRF Star IV, US GlobalSat, FL, USA).
The endurance of this rig is limited by the battery of the laptop used
for recording and for monitoring the camera view while walking
with the rig.

\section{Forest environment dataset}
\begin{figure}
	\centering
	\includegraphics[width=3in]{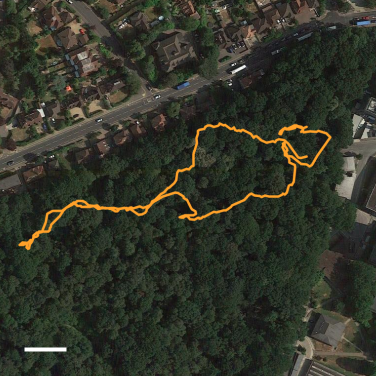}
	\caption{\textbf{Sample path for a data collection run.}
          Trajectory (orange) from GPS meta data of data collection
          Run~1 (see Tab.~1) overlaid on aerial view to illustrate
          the recording environment in the Southampton Common. The
          white scale bar corresponds to a distance of 30~m. For all
          image frames of all runs the GPS metadata is included with
          the dataset.
          Google Maps background: Imagery\copyright 2020 Getmapping
          plc, Infoterra Ltd \& Bluesky, Maxar Technologies; permitted
          use.}
          \label{fig:path}
          %
          %
          %
          %
\end{figure}

The data for our forest environment dataset was collected in woodland
areas (Fig.~\ref{fig:path}) of the $1.48~\mbox{km}^2$ Southampton Common (Hampshire, UK).

\begin{figure*}[t!]
  \centering
  \includegraphics[width=.75\textwidth]{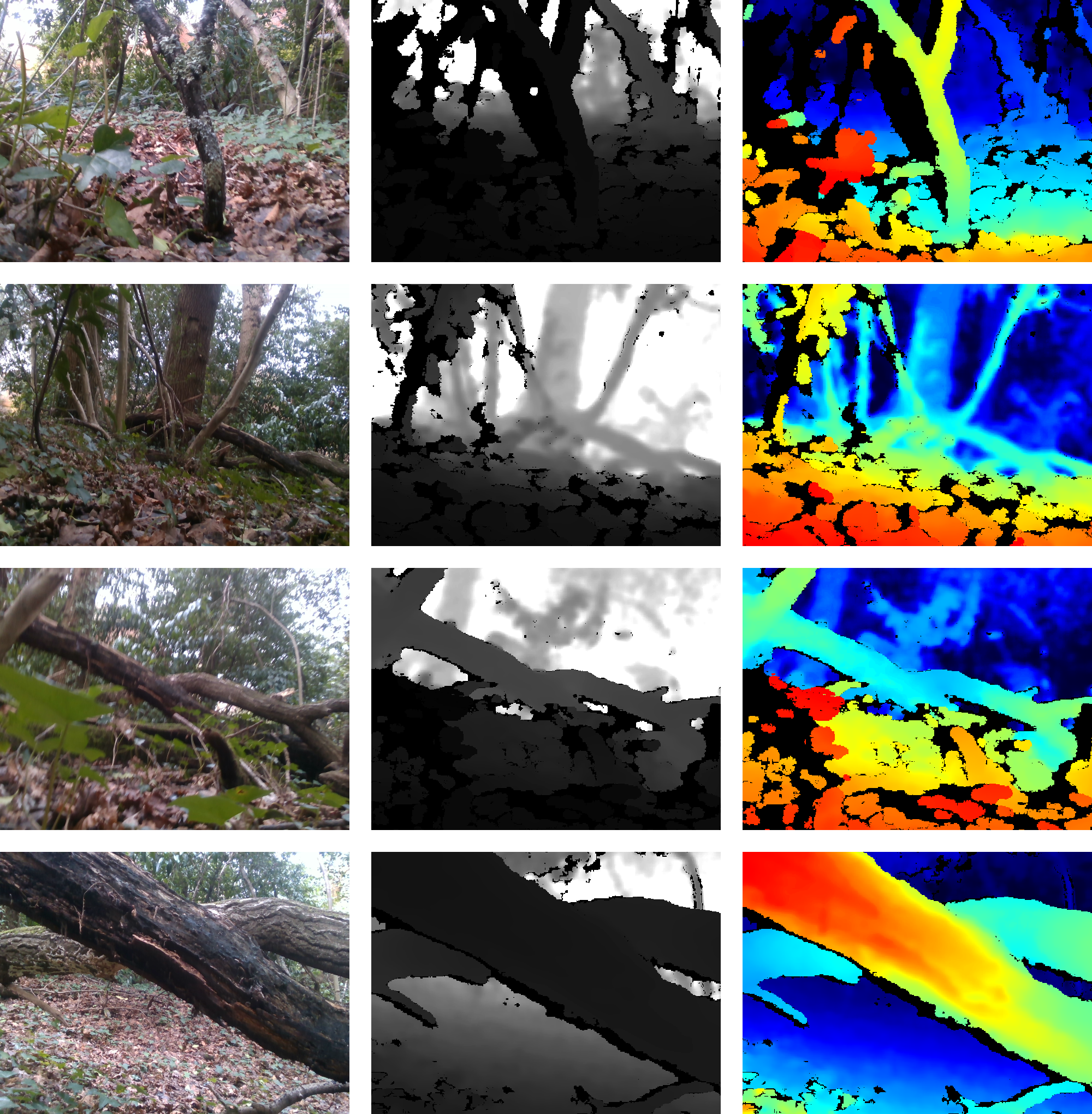}
  \caption{\textbf{Sample scenes from the forest environment dataset.} A diverse set of scenes in RGB (left), and the aligned depth in grayscale (middle) and color (right), were recorded in the forest. In grayscale, lighter pixels indicate regions further from the camera, and white pixels are out of range. The gradients in depth are better illustrated in color, with warmer colored pixels indicating regions closer to the camera. In both color schemes, black pixels indicate the depth could not be estimated.}
  \label{fig:example-frames}
\end{figure*}

The data collection rig was pushed through the forest area in the
Southampton Common in five separate runs during different times of day
and different weather conditions to sample variations in lighting.
Table~\ref{tab:env} shows the recording conditions, where the
luminosity values are normalised to range from 0.0 to 1.0 in arbitrary
units and give the average over the luminosity of all frames (see next
section) in the run.
\begin{table}
  \caption{Forest environment recording conditions.
    Luminosity in arbitrary units, see text for details.}
  \label{tab:env}
  \begin{center}
    \begin{tabular}{m{0.1\linewidth}m{0.13\linewidth}m{0.2\linewidth}m{0.13\linewidth}m{0.15\linewidth}}
      \toprule
      Dataset index & Weather condition      &Time of day & Number of images recorded & Mean luminosity \\
      \midrule
      Run 1 & Partly sunny  & Midday & 27,333 & 0.41\\
      Run 2 & Scattered clouds  & Midday & 33,194 & 0.41\\
      Run 3 & Cloudy, light rain  & Evening & 20,328 & 0.31\\
      Run 4 & Sunny  & Afternoon & 17,499 & 0.38\\
      Run 5 & Mostly clear & Morning & 36,331 & 0.37\\
      \bottomrule
    \end{tabular}
  \end{center}
\end{table}
Sample forest scenes from the runs are shown in
Fig.~\ref{fig:example-frames}. For each run in the forest the
following data was recorded from the sensor platform: (i) RGB and
depth images from the camera, (ii) Linear acceleration and angular
velocity from the six degree-of-freedom IMU of the camera, cf.
Fig.~\ref{fig:platform} for axes orientation, (iii) rotary encoder
counts, and (iv) GPS position of the platform.

The data from the rotary encoder and IMU streams were time
synchronized with the recorded images from the camera at 30 frames per
second, and recorded at the same rate. The GPS location data was also
synchronized with the camera feed, and recorded once per second.
Recorded image data was stored lossless in 8-bit PNG file-format at
$640 \times 480$ pixel resolution. Data from the IMU, rotary encoder
and GPS sensors were stored for ease of access as comma-separated
values in a plain text file. Our full forest
data-set comprises over 134000 RGB/depth image pairs with concomitant
metadata. A convenient subset containing about 9700
aligned RGB and depth image pairs with the corresponding time synchronized
IMU, rotary encoder, and GPS information is available online \url{DOI:
  10.5281/zenodo.3945526} under Creative Commons
Attribution 4.0 International Public License.

\section{Quality of our forest environment dataset}

To assess the image quality of the depth data in our forest
environment dataset we consider, (i) the \textit{fill rate}, which is
the percentage of the depth image containing pixels with a valid
estimated depth value, (ii) the \textit{depth accuracy} using ground truth
data, and (iii) the \textit{image perspective} that can be determined by camera orientation.

\noindent\textbf{Fill rate of depth images:}
The depth camera uses stereo vision to calculate depth, but augments
this technique by projecting with an infra-red laser a dot pattern
into the scene. This process should be reasonably robust against
camera motion, but could potentially be susceptible to illumination
levels of the scene.
For our analysis, the instantaneous velocity and acceleration of the
mobile sensor platform was estimated using the rotatory encoders data.
As a proxy for actual illumination measurements we calculate a scalar
luminosity (perceived brightness) value from the color channels of the
RGB pixels and averaged over all pixels in the image to arrive at
the normalised luminosity of the frame (arbitrary units).

\begin{figure}[htbp]
  \centering
  \includegraphics[width=3.2in]{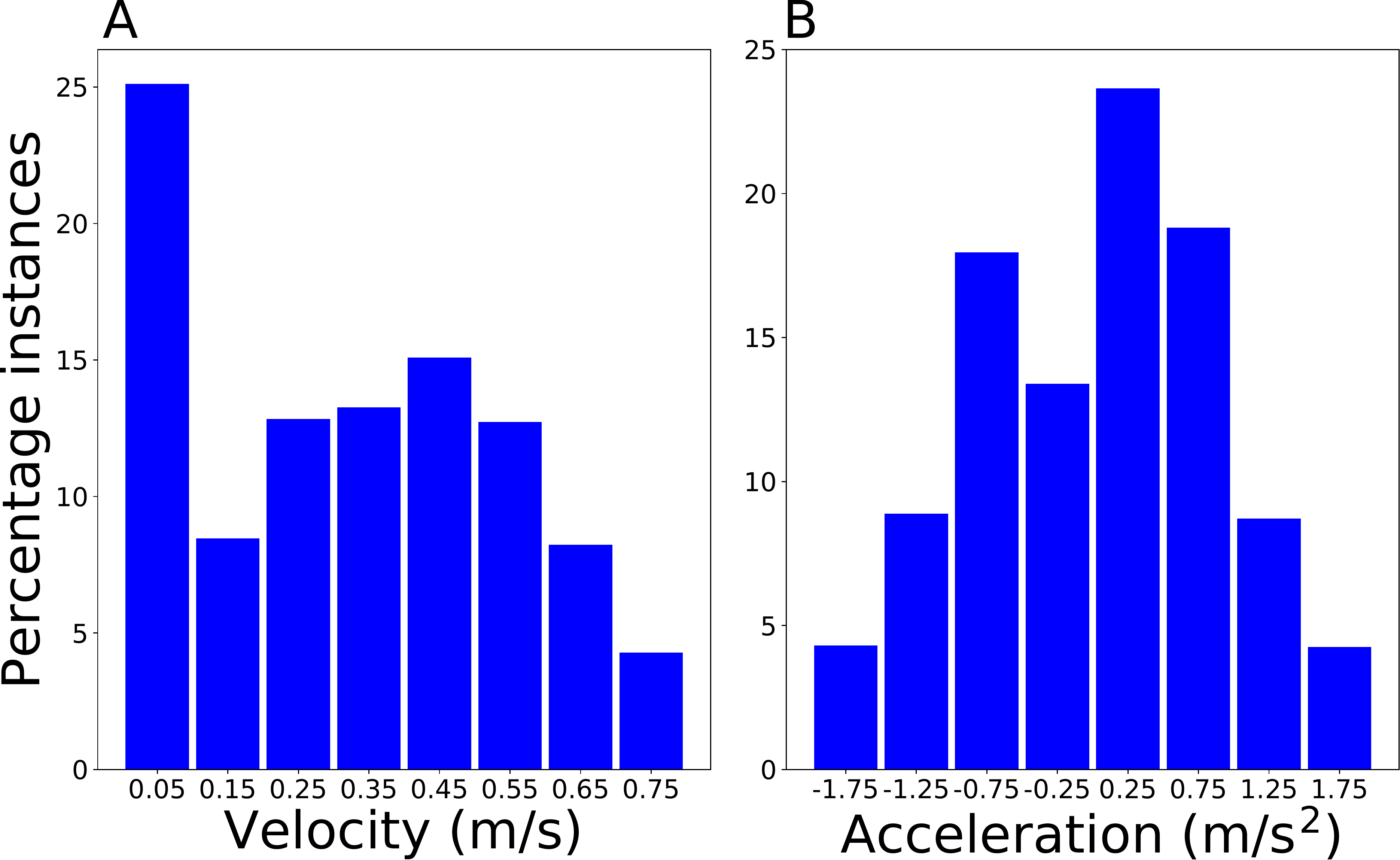}
  \caption{\textbf{Velocity and acceleration during recording.} The
    linear velocity and acceleration of the mobile sensor platform in
    the forward direction, while being pushed through the forest. Data
    for the distribution was aggregated across all five runs of the
    dataset. Instantaneous velocity and acceleration were estimated
    from the rotatory encoder data.}
  \label{fig:platform-speed}
\end{figure}

\begin{figure}
	\centering
	\includegraphics[width=3.2in]{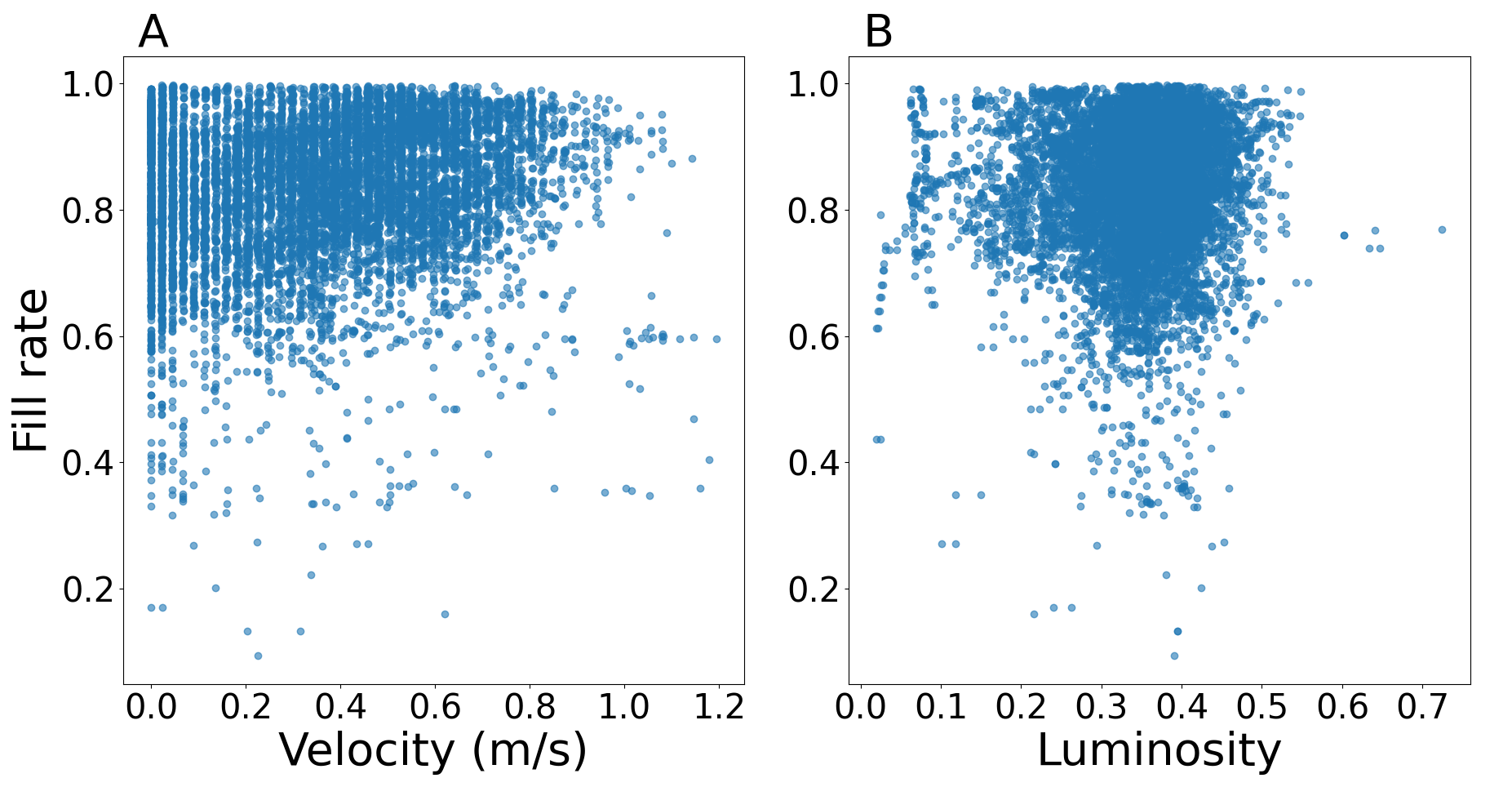}
	\caption{\textbf{Sampled fill rate}. Changes in velocity (A)
          and lighting (B) do not affect the fill rate over the range
          encountered in the five runs. For clarity the panels show
          data for 1000 frames randomly selected from all five runs.}
	\label{fig:vlf}
\end{figure}

The recording rig was pushed at speeds comparable to what we expect
for portable robots in the forest environment
(Fig.~\ref{fig:platform-speed}). We found that over this speed range
the fill rate is not affected by the velocity of the camera, as seen
in Fig.~\ref{fig:vlf}A. Similarly, the fill rate is not affected by the
luminosity of the scene (Fig.~\ref{fig:vlf}B) and generally across the
luminosity and velocity range tested the camera achieves a reasonably
high fill rate (mean $0.84\pm0.11$ SD across all depth images from all five runs).

\begin{figure}
  \centering
  \includegraphics[width=2.5in]{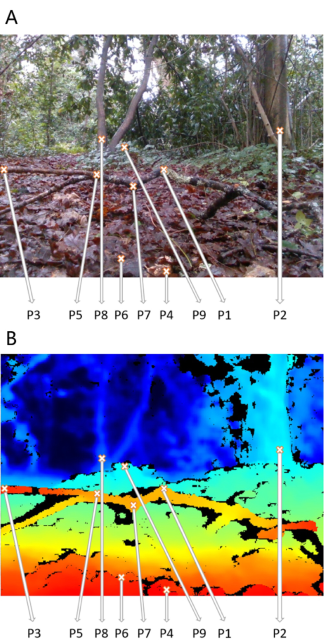}
  \caption{\textbf{Position of sampled points for accuracy of depth images.} Nine points at varying depth and positions were sampled from a typical forest scene. Points 1, 3, 5 and 7 are on a fallen tree brach, points 4 and 6 are part of the forest floor, particularly close to the camera, and points 2, 8 and 9 are located on tree trunks close to the ground. The points 4 and 8, are nearest to and furtherst from the camera, respectively.}
  \label{fig:depth-points}
\end{figure}

\noindent\textbf{Accuracy of depth images:} To evaluate the accuracy
of the depth images, we established ground truth depth measurements with a Zamo
Digital distance meter (Bosch, Germany; maximum range $20$~m, accuracy
$\pm2$~mm). For ground truth measurements nine points at
varying depths in a typical forest scene were considered. The selected
points, depicted in Fig.~\ref{fig:depth-points}, were located on the
forest floor, on fallen leaves, fallen tree branches, and low on tree trunks.
Ground truth measurements were replicated thrice for each of the nine
selected points. An offset of 4.2mm was added to values returned by
the ground truth sensor to account for differences in its incident
position and that of our depth camera. To account for the divergence
of the laser from the ground truth sensor, depth estimates with our
depth camera were averaged over $7 \times 7$ pixels at the laser spot.
Two independent depth-images were used to have a replication of the depth measurement from the camera.
\begin{figure}
  \centering
  \includegraphics[width=3.5in]{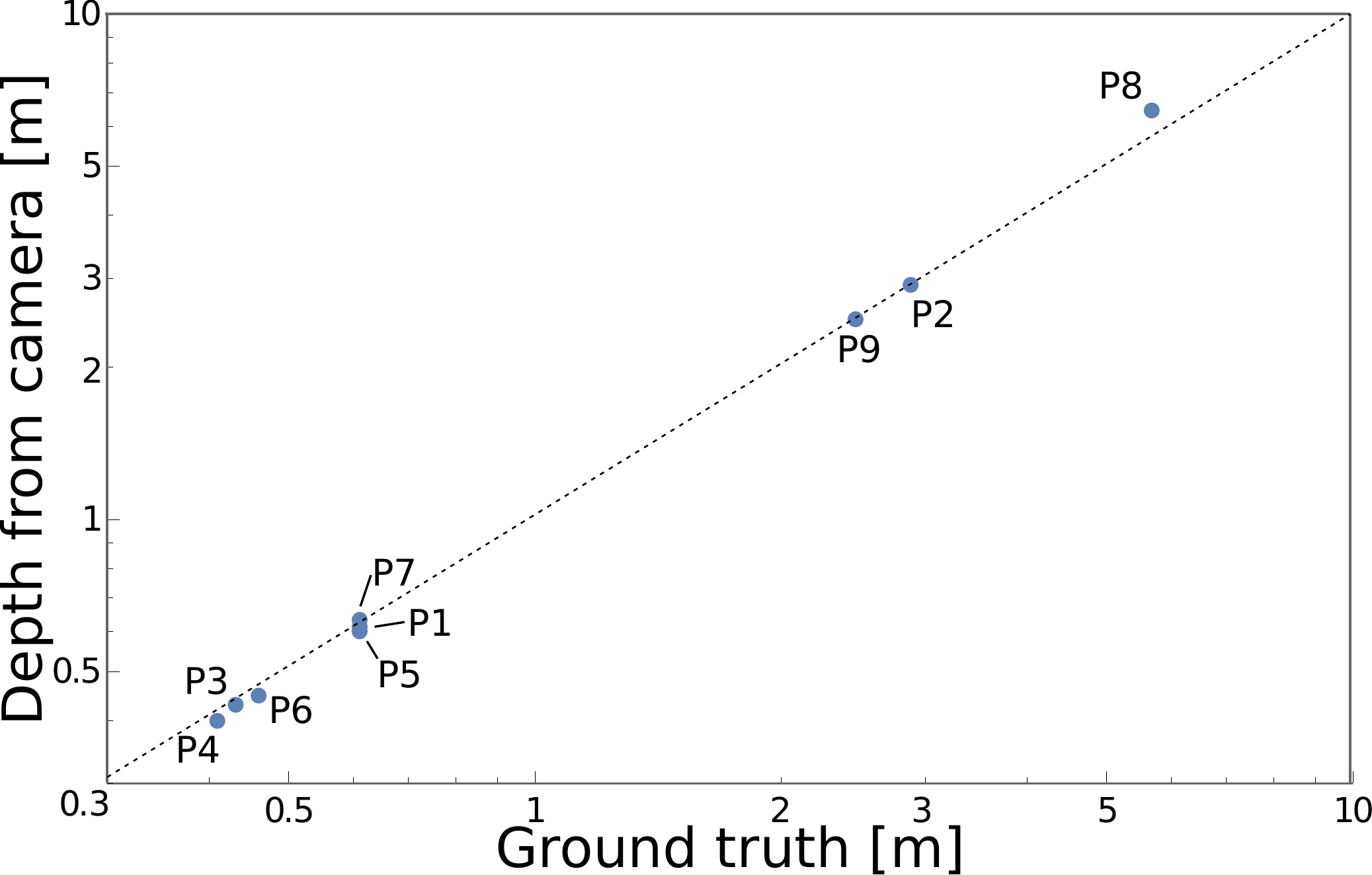}
  \caption{\textbf{Accuracy of the depth data.} The accuracy of the
    depth information for nine sample points, P1 to P9. Ground truth measurements were averaged over three replicates. Depth image data was averaged over $7 \times 7$ pixels at the laser spot and over two replicates. Points on the diagonal dotted line indicate depth estimates identical to ground truth measurements.}
  \label{fig:depth-error}
\end{figure}
As can be seen in Fig.~\ref{fig:depth-error}, the information from the
camera corresponds well with the ground truth measurements (see
Fig.~\ref{fig:depth-error}). Across all sampled points P1 to P9, the mean error was less than
$4\%$. The highest deviation of $12\%$ was at point P8, which was positioned furthest from the camera.

\begin{figure}
	\centering
	\includegraphics[width=2.8in]{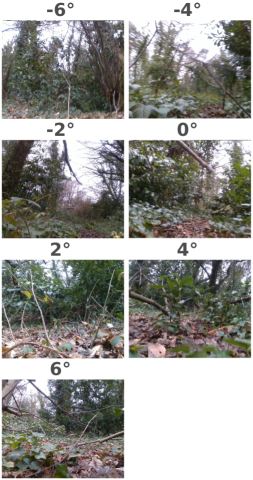}
	\caption{\textbf{Image instances for different pitch angles.}
          Perspectives ranging from -6$^{\circ}$ to 6$^{\circ}$ camera pitch angle.}
	\label{fig:filter}
\end{figure}

\noindent\textbf{Image Perspective:}
Approximately 15--25\% of the image frames in each video were taken
with the camera tilted upwards and do not include the ground in the
view. For our purpose of training depths estimating neuronal networks,
such frames are helpful, because they do not have the direct
correlation between distance and height (y-axis position) that is
otherwise common. In applications where frames without ground in view are
undesirable, such frames can be excluded as follows. First the raw
accelerometer and gyroscope data from the IMU as metadata for each
frame, is fused to arrive at an absolute orientation for the camera.
Positive pitch values indicate a downward perspective, a threshold can
be set to discard frames in which the camera is tilted backwards.
After low-pass filtering the pitch angle of the camera
\cite{press1990savitzky}, frames without the ground in view can be
discarded by filtering out frames with an angle below -4  degrees.
Sample images at different camera pitch are shown in
Fig.~\ref{fig:filter}. For convenience the pitch values---in a
addition to the raw IMU data---have been included in the metadata of
the forest depth dataset.



\section{Conclusions}

\begin{figure}
	\centering
	\includegraphics[width=3.2in]{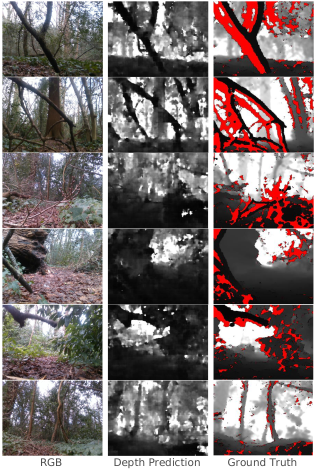}
	\caption{
          \textbf{Sample results for depth estimation.}
          Results from a U-net \cite{ronneberger2015u} trained with
          8204 RGB-depth image pairs. In the depth images (right column),
          pixels without vaild depth information are indicated in red.
          The U-net receives the RGB image in the left column as input
          and provides the depth estimation shown in the center
          column. The predicted depth map can be compared to the recorded
          depth image in the right column.
        }
	\label{fig:depth}
\end{figure}

An off-road forest depth map dataset has been collated to support the development of computer vision modules for portable robots that operate in forests. Accordingly, it is recorded from a low viewpoint and with close-up views of obstacles such as fallen tree branches and shrubs. The data set is of sufficient size to train modern neuronal networks and is provided with metadata, that can, for example, be used to filter the frames by camera orientation. We created this dataset with the primary aim to develop robots of sufficiently low cost that sparse robot swarms \cite{Tarapore2020SparseSwarms} will become feasible. In this context, it is of interest to replace depth cameras with estimated depth information from RGB images. In ongoing work we are using the dataset to develop such depth prediction models particularly targeted at low capability embedded computers; a sample of what can be expected is shown in Fig.~\ref{fig:depth}. We believe that the dataset is a first step to fill the gap of out-door datasets for small robots and that it will be of use to the community. For example, with the steering (rotary encoder) information available in the metadata, it may be possible to use the dataset to train an autonomous guidance system. Hopefully this contribution will stimulate computer vision research in the nascent and challenging field of forest robotics.



\bibliographystyle{IEEEtran}
\bibliography{IEEEabrv,myref}

\end{document}